\documentclass[conference]{IEEEtran}
\IEEEoverridecommandlockouts
\usepackage{amsmath,amssymb,amsfonts}
\usepackage{algorithmic}
\usepackage{graphicx}
\usepackage{textcomp}
\usepackage{xcolor}
\usepackage{color}
\usepackage{subcaption}
\usepackage{soul}

\usepackage[backend=biber,style=ieee,minnames=1,maxnames=1,dashed=False]{biblatex}
\AtBeginBibliography{\footnotesize}

\usepackage[left=0.65in, right=0.65in, bottom=1in, top=0.8in]{geometry}

\def\BibTeX{{\rm B\kern-.05em{\sc i\kern-.025em b}\kern-.08em
    T\kern-.1667em\lower.7ex\hbox{E}\kern-.125emX}}

\renewcommand{\baselinestretch}{0.82}

\begin{document}

\title{Online Meta-Learning Channel Autoencoder for Dynamic End-to-end Physical Layer Optimization\\
\thanks{This material is based upon work supported by the National Science Foundation under Grant Numbers  CNS-2202972, CNS-2232048, and CNS-2318726, and the AFRL Summer Faculty Fellowship Program (SFFP) under contract number FA8750-20-3-1003.}
\thanks{DISTRIBUTION STATEMENT A: Approved for Public Release; distribution unlimited AFRL-2024-2359 on April 30, 2024.}
}

\author{
	\IEEEauthorblockN{
	Ali Owfi\IEEEauthorrefmark{1}, 
        Jonathan Ashdown \IEEEauthorrefmark{2},
        Kurt Turck \IEEEauthorrefmark{2},
        Fatemeh Afghah\IEEEauthorrefmark{1}}

    \IEEEauthorblockA{\IEEEauthorrefmark{1}Holcombe Department of Electrical and Computer Engineering, Clemson University, Clemson, SC, USA \\
        Emails: \{aowfi,  
        fafghah\}@clemson.edu}

    \IEEEauthorblockA{\IEEEauthorrefmark{2}Air Force Research Laboratory,  Rome, NY, USA \\
	 Emails: \{jonathan.ashdown,kurt.turck\}@us.af.mil}

}

\maketitle

\begin{abstract}

Channel Autoencoders (CAEs) have shown significant potential in optimizing the physical layer of a wireless communication system for a specific channel through joint end-to-end training. However, the practical implementation of CAEs faces several challenges, particularly in realistic and dynamic scenarios. Channels in communication systems are dynamic and change with time. Still, most proposed CAE designs assume stationary scenarios, meaning they are trained and tested for only one channel realization without regard for the dynamic nature of wireless communication systems. Moreover, conventional CAEs are designed based on the assumption of having access to a large number of pilot signals, which act as training samples in the context of CAEs. However, in real-world applications, it is not feasible for a CAE operating in real-time to acquire large amounts of training samples for each new channel realization. Hence, the CAE has to be deployable in few-shot learning scenarios where only limited training samples are available. Furthermore, most proposed conventional CAEs lack fast adaptability to new channel realizations, which becomes more pronounced when dealing with a limited number of pilots. To address these challenges, this paper proposes the Online Meta Learning channel AE (OML-CAE) framework for few-shot CAE scenarios with dynamic channels. The OML-CAE framework enhances adaptability to varying channel conditions in an online manner, allowing for dynamic adjustments in response to evolving communication scenarios. Moreover, it can adapt to new channel conditions using only a few pilots, drastically increasing pilot efficiency and making the CAE design feasible in realistic scenarios.

\end{abstract}

\begin{IEEEkeywords}
Adaptive modulation, Adaptive physical layer, Channel Autoencoder, Online Meta learning, Few-shot learning.
\end{IEEEkeywords}

\section{Introduction}

Amidst the rapid progress of artificial intelligence, Deep Learning (DL) and Machine Learning (ML) methods have found applications in critical wireless communication tasks, including device and modulation identification, channel 
estimation, beamforming, and interference mitigation \cite{Qin}.  
Among DL-based methods in wireless communication applications, utilization of AEs in physical layer optimization and design has gained significant traction due to its capability to offer end-to-end design solutions and enable joint coding and modulation, as this end-to-end optimization potentially increases overall performance compared to individually optimized blocks in traditional physical layer designs.
Autoencoders (AE) are a class of unsupervised machine learning models known for their ability to learn efficient data representations. They consist of two main components: an encoder, which compresses input data into a latent representation, usually in a lower dimension, and a decoder, which reconstructs the original data from the latent representation. AEs find applications in generative models \cite{huang2019variational}, denoising\cite{owfi2023autoencoder}, and representation learning \cite{chen2023context}, where they help extract meaningful features from raw data.

\begin{figure}[htp]
\begin{center} 
\vspace{-10 pt}
\includegraphics[width=0.4\textwidth]{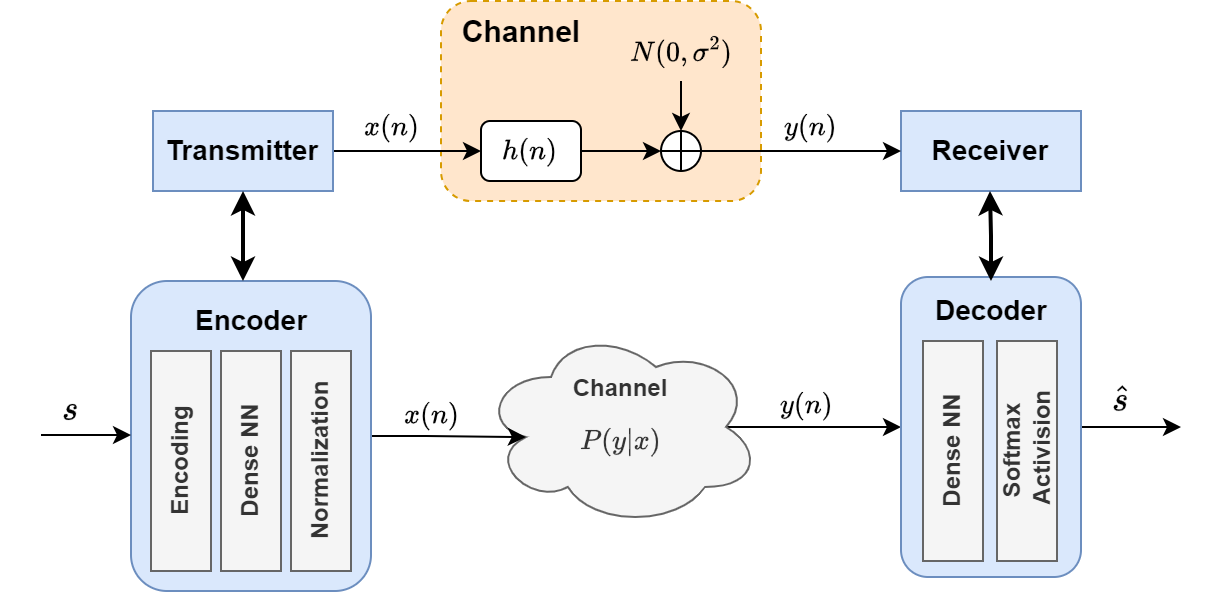}
\caption{\small{Block diagram of the general CAE design.}} 
\label{fif:ae_channel} 
\end{center} 
\vspace{-5pt}
\end{figure}

The encoding and decoding architecture of AEs, resembling that of communication systems, inspired the novel Channel Autoencoder (CAE) concept, depicted in Figure \ref{fif:ae_channel}. In CAEs, the conventional roles of transmitter and receiver are substituted with an encoder and decoder, respectively. In contrast to conventional communication models that rely on multiple individual functional blocks optimized separately, CAE designs offer a holistic, data-driven, and end-to-end learning-based optimization approach. Many studies have demonstrated that this end-to-end joint optimization, combined with the inherent capacity of AEs to optimize complex systems, leads to hand-tailored, optimized physical layer settings that outperform standard conventional physical layer configurations, resulting in higher performance in various communication scenarios \cite{wu2020deep,zou2021channel,felix2018ofdm}.

\begin{figure}
\vspace{-5pt}
\centering
\begin{subfigure}[b]{0.4\textwidth}
   \includegraphics[width=1\linewidth]{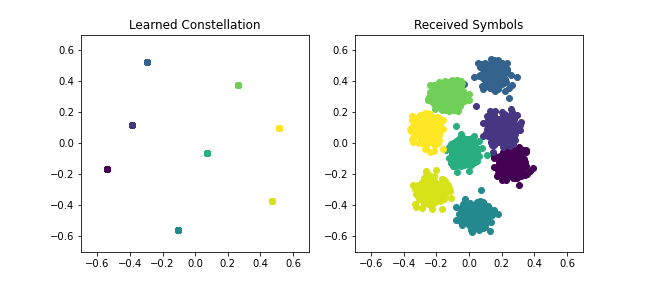}
\end{subfigure}

\begin{subfigure}[b]{0.4\textwidth}
   \includegraphics[width=1\linewidth]{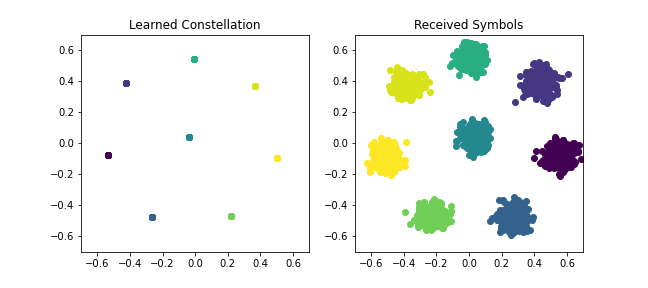}
\end{subfigure}
\caption{\small{Learned constellation (left) and received symbols (right) under two different fading channels.}}
\label{fig:constellation}
\vspace{-5pt}
\end{figure}

Despite the promising advantages of CAE design and its recent advancements, several critical challenges must be addressed before its practical implementation in real-time communication systems. Figure \ref{fig:constellation} depicts a CAE's learned optimal constellation designs for two different Rayleigh fading channels. As the figure shows, the solution for one channel cannot be directly applied to another. Conventional CAEs often require training from scratch for optimal performance under each distinct channel setting. This process is undesirable as it demands substantial computational resources and a large amount of new data, which may not be feasible in real-time scenarios. Moreover, the dynamic nature of communication channels underscores the need for a well-generalized or highly adaptable solution for physical layers, further highlighting the existing limitations in the conventional CAE design's adaptability to new channels. 

Furthermore, in CAEs, pilot signals are considered the means of relaying ground truth labels from the transmitter to the receiver, effectively acting as training samples for the CAE. As pilot signals are much smaller than the actual transmitted data, practical implementation of CAE will probably face a few-shot scenario, as the number of pilots acting as training samples could be limited. However, the assumption of few-shot learning for CAE is absent from most current studies.

Meta-learning offers a promising solution to address the lack of adaptability in current CAE designs and few-shot learning scenarios. Meta-learning, an ML technique, enables models to learn how to learn, allowing them to adapt quickly to new tasks or environments with a significantly smaller new dataset and, subsequently, computation. In the context of CAE design, meta-learning can be employed to develop models that can efficiently adapt to varying channel conditions using a few pilots and without requiring extensive retraining from scratch. 

Furthermore, a common underlying assumption in the recently proposed CAEs is the availability of all training data from the beginning, meaning that the training is done offline \cite{o2017physical,jiang2019turbo,zou2021channel}. This assumption does not accurately represent real-time scenarios, where data from a communication channel is typically provided sequentially. This sequential data arrival pattern aligns more closely with the online learning paradigm, which is currently mostly absent in CAE studies.

To provide an adaptable CAE solution to more realistic and dynamic channel settings with few-shot constraints, we propose an Online-Meta-learning Channel Autoencoder framework named OML-CAE. \textit{To the best of our knowledge, we are the first to develop an Online-Meta-Learning framework for CAE under few-shot learning constraints for online dynamic fading channels.}

The main contributions of this paper are:
\begin{itemize}
    \item Introducing an online meta-learning framework for CAEs,  enhancing adaptability and real-time learning capabilities, offering a more practical CAE design for realistic dynamic scenarios.
    \item Redefining the CAE setup to address practical limitations by formulating an \underline{online} scenario with varying fading channels. The proposed OML-CAE is gradually trained in an online manner as new fading channels are encountered, as opposed to conventional CAEs that are trained offline and under only one channel realization.
    
    \item Solving the CAE task as a few-shot learning scenario by utilizing meta-learning. In the context of CAEs, pilots serve as the training samples. However, in practical scenarios, obtaining a vast number of pilots for each channel realization is often impractical. Unlike conventional CAEs, which demand ample training samples, our proposed approach, OML-CAE, is capable of training with just a few samples from a single channel realization.
    
    \item Reducing pilot overhead by lowering the number of training samples required to reach optimal performance. In communication systems, lowering pilot overhead is important to  increase efficiency, data rates, and transmission capacity. The proposed OML-CAE delivers the high accuracy of what CAE designs offer compared to the traditional physical layer methods but requires drastically fewer pilots to be trained than conventional CAE designs.

\end{itemize}


\section{Related Work}


The idea of DL-based optimized physical layer system has gained a lot of attention in recent years. The authors in \cite{o2017introduction}, proposed the novel idea of modeling a simple communication system as a channel AE to jointly optimize the transmitter and the receiver for a two-user AWGN channel. The proposed design introduced a new data-driven paradigm to physical layer design, which was further extended in subsequent works. While the original work considered a two-user interference channel, the authors in \cite{wu2020deep} applied the channel AE design to the multi-user interference channel. The channel AE has also been studied under fading channels \cite{zhu2019joint}, demonstrating higher performance compared to traditional modulation schemes. In \cite{dorner2018deep}, the authors validated the performance of the channel AE design, which had only been confirmed through simulations before, in over-the-air transmissions. Additionally, they introduced a frame synchronization scheme to the channel AE to address the challenge of receiver synchronization.

Wireless communication systems are characterized by their highly dynamic nature, where channel conditions are constantly changing. This dynamic nature poses a significant challenge for DL models applied to wireless systems, as they usually require substantial amounts of training data to effectively capture and adapt to these variations. As a solution, meta-learning has been applied to various wireless communication applications in the past years to enable rapid adaptability of DL models to changing environments, such as in wireless device localization \cite{owfi2023metalocal}, signal identification\cite{owfi2023meta}, 
 and demodulation \cite{park2020learningdemod}. In the context of CAE, limited work has been done for adaptability to unseen channel realizations in dynamic scenarios and, more specifically, integration of meta-learning for CAE design. The authors in \cite{park2020meta} addressed the issue of adaptability and proposed meta-learning for channel AE with the aim of faster adaptation to new fading channel realizations. However, they did not consider the probable constraint of limited training samples for new channel realizations. Furthermore, the study was not conducted in an online setting.

 The authors in \cite{park2020end} proposed the utilization of online meta-learning for dynamic fading channels. Based on the code they have provided, the meta-learning process is not done in an online manner, rather, meta-learning is performed offline on a selection of different fading channels, and then the adaptation phase is done online as new fading channels are encountered. This acts as a meta-pre-training which does lead to better adaptation than a conventional CAE. However, since the meta-training isn't conducted online, this framework's accuracy will diminish over time. This occurs as the initial offline meta-training relies on a fixed set of fading channels, which can become outdated. Moreover, while they considered scenarios with a limited number of pilot blocks, the number of data samples per symbol is not limited. This means long pilots are used for training. Therefore, even the scenario referred to as limited pilot blocks uses plentiful training samples and differs from a few-shot learning scenario.

While all mentioned meta-learning-based studies have provided valuable insight into the integration of Meta-Learning in CAE design, they either are not online or do not consider few-shot scenarios. To the best of our knowledge, the proposed OML-CAE framework in this paper is the first method to address the adaptability of CAEs to dynamic fading channels in an online scenario under the few-shot learning constraint.

\section{Methodology}

\subsection{Online Meta Learning}

Meta-learning, or "learning to learn," is a subfield in machine learning that focuses on algorithms that quickly adapt to new tasks using the accumulated knowledge in learning previously seen tasks. Unlike conventional DL methods, meta-learning methods aim to learn the learning process, enabling swift adaptation to new unseen tasks with a smaller training dataset and a more efficient adaptation process.

MAML (Model-Agnostic Meta-Learning)\cite{finn2017model} is a prominent algorithm in meta-learning. Being model-agnostic, it can be applied to various types of models, providing a versatile framework for meta-learning across different domains. Formally, MAML considers an inner model $f$ with parameters $\theta$ denoted by $f_\theta$. The primary goal of MAML is to find an optimized $\theta$ for the inner model from which, adaptation to new tasks can occur rapidly. In MAML, tasks are divided into meta-training and meta-testing tasks, each task having its own objective, support set (training set), and query set (test set). MAML consists of two main loops: the inner loop and the outer loop. During the inner loop or the adaptation phase, the meta-learning model adapts to a selected task $T_i$ by training on its support set and updating $\theta$ to compute the task-specific optimized parameters $\theta'_i$:

\begin{equation}
    \theta_{i}^{'} = \theta - \alpha\Delta_{\theta}\mathcal{L}_{\mathcal{T}_i}(f_{\theta})
\end{equation}

During the outer loop, the inner model's initial parameters $\theta$ are updated based on the task-specific losses calculated from the task-specific optimized parameters $\theta'_i$ computed in the inner loop:

\begin{equation}
\theta \leftarrow \theta-\beta \nabla_{\theta} \sum_{\mathcal{T}_{i} \sim p(\mathcal{T})} \mathcal{L}_{\mathcal{T}_{i}}\left(f_{\theta_{i}^{\prime}}\right)
\label{eq:outer_update}
\end{equation}
where $\beta$ is a hyper-parameter known as \emph{meta-step size}. 
The outer loop which results in updated initial parameters $\theta$ for the inner model is only executed on the meta training tasks. For the meta-testing tasks, which are not supposed to contribute to the meta-learning process, only the inner loop is executed so that the performance of adaptation to new unseen tasks can be evaluated. More details about MAML can be found in the original paper\cite{finn2017model}.

In the original MAML algorithm, it is considered that meta-training tasks are all available from the beginning. Hence, the meta-learning process can be considered to be done offline, and when an unseen task is sampled, the meta-learned model can adapt to it through an adaptation phase. To have an online meta-learning scenario, we can consider a case where tasks are encountered sequentially in time. In this scenario, at sequence $i$, we can treat all tasks encountered in the previous sequence $T_j$  $j<i$, stored in a task buffer, as meta-training tasks to be used for the meta-training process, and the current task $T_i$ as the meta testing task. Naturally, at the start of the sequence, the task buffer would be empty, and meta-training tasks would be accumulated as sequences progress.

\subsection{Problem Statement}

We consider a channel AE setup in a single-link communication system shown in Fig. \ref{fif:ae_channel}. The AE setup consists of three components: an encoder
$f_{T}(.)$ with parameters $\theta_T$, a channel $c (.)$, and a decoder $f_{R}(.)$ with parameters $\theta_R$. Given a message $m \in \{1, ...,  2^k\}$ where $k$ represents the number of bits, the encoder takes \textbf{s}, one hot vector of message $m$, and outputs transmitted vector \textbf{x} of size $2n$, where $n$ is the number of complex channel uses. In this problem formulation, the number of channel uses specifies the number of independent channels that can be used for simultaneously transmitting different segments of the message, similar to the number of sub-carriers in an OFDM system.  As depicted in Fig. \ref{fif:ae_channel}, the encoder goes through a normalization layer to ensure the power constraint of $E(\textbf{x}^2) = 1$ is met.
The transmitted signal $\textbf{x}$ is then transmitted over the channel $c(.)$ where the received signal is created by: $\textbf{y} = \textbf{h}\textbf{x} + \textbf{n}$, where $\textbf{h}$ represents the channel vector of a fading channel and $n \sim N(0, \sigma^2)$ represents Gaussian i.i.d. noise. The decoder receives the received signal $\textbf{y}$ as its input and provides and outputs probability vector $\hat{\textbf{s}} = \{p(m_1|\textbf{y}), ..., p(m_{2^k}|\textbf{y})\}$ of size $2^k$. The loss is then calculated based on $argmax(\hat{\textbf{s}}) \neq \textbf{s}$. The autoencoder parameters $\theta_T$ and $\theta_R$ can then be optimized using SGD with respect to the calculated loss. 



The above channel AE setup has demonstrated effective performance when confronted with a fixed fading channel and ample training data. However, the proposed conventional channel AE is expected to fail in dynamic and more realistic scenarios. We consider an online scenario with $n$ sequences, wherein in each sequence, a new fading channel is realized from an autoregressive Rayleigh block fading process. More formally, in time sequence $i$, the channel $h_i$ is realized based on $h_i = \rho h_{i-1} + \sqrt{1-\rho^2}h'$, where $h_{i-1}$ is the Rayleigh fading channel realization in the previous sequence, $h'$ is a Rayleigh fading channel realization acting as the innovation term in the autoregressive process, and $\rho$ is a correlation coefficient \cite{park2020end}. As observed in Fig. \ref{fig:constellation}, for each distinctive channel the optimized learned constellation differs, and the learned constellation for channel $h_i$ will not perform well for channel $h_j$ without further training.

Moreover, in the conventional channel AE setup, limitations on the availability of training data are commonly disregarded. To calculate the loss for the channel AE framework, we have to have the ground truth transmitted message labels at the decoder. We can consider that the ground truth labels can be transmitted to the receiver using pilot signals, but nevertheless, pilot signals are much smaller in size than the actual transmitted data. To this end, if the channel AE framework is going to be practically utilized, it is more than likely that channel AE will face a few-shot learning scenario, where the number of available training data is limited to just a few samples per message. This issue further highlights the shortcomings of the conventional channel AE design.

\subsection{Online Meta-Learning Channel AE (OML-CAE)}
To address the limitations faced by a CAE in realistic and dynamic scenarios—specifically, the lack of adaptability, scarcity of training samples, and the need for real-time online performance—we propose an Online Meta Learning Channel AE framework, named OML-CAE. As mentioned earlier, meta-learning stands out as a powerful tool for enhancing generalization and facilitating rapid adaptation, particularly in few-shot learning scenarios. The challenges that meta-learning effectively tackles align well with the requirements of a Channel AE setup, where the model must quickly adapt to new channel realizations with only a small set of training samples which are transmitted as pilot signals. Furthermore, we propose incorporating meta-learning for channel AE in an online manner, addressing the need for real-time performance channel AEs.

\begin{figure}[htp]
\begin{center} 
\vspace{-5pt}
\includegraphics[width=0.49\textwidth]{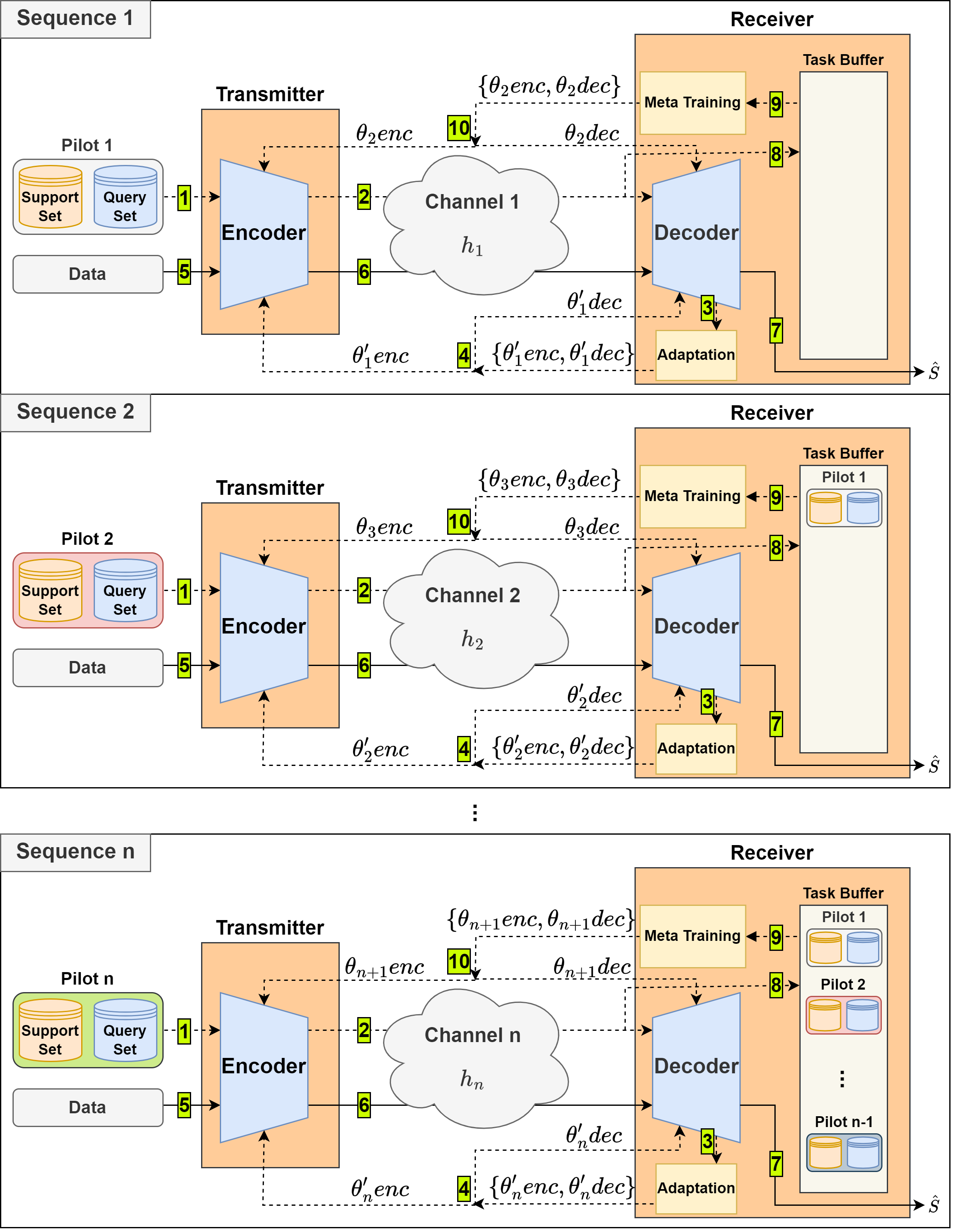}
\caption{\small{Overview of the OML-CAE framework. Steps related to the pilot and data are shown with dashed and solid lines respectively. At sequence i, with the fading channel realization i The encoder and the decoder start with initial parameters $\theta_{i}enc$ and $\theta_{i}dec$. In steps 1 and 2, the pilot is encoded and transmitted through the channel. In step 3, the transmitted pilot signals are decoded, and a loss is calculated based on the few labeled samples available in the pilot. In step 4, adaptation (inner loop) is performed and task-specific optimized parameters for the encoder ($\theta'_ienc$) and the decoder ($\theta'_idec$) are obtained. After the CAE is adapted for the given channel, the data is transmitted and decoded through steps 5, 6, and 7. No updates are done in these steps and the output only determines the model's accuracy for the given fading channel. In step 8, the received pilot is added to the task buffer. In the first sequence, the task buffer is empty and the task buffer is sequentially filled as the model observes new channel realizations. In steps 9 and 10, meta-training is performed using all the pilots stored in the task buffer. Through the meta-training, the initial parameters for the encoder ($\theta_{i+1}enc$) and the decoder ($\theta_{i+1}dec$) are updated which will be used in the next sequence when the fading channel changes.}}
\label{fif:online} 
\end{center} 
\vspace{-5pt}
\end{figure}

The overview of the proposed OML-CAE framework is presented in Fig. \ref{fif:online}. The process of optimizing the transmitter (encoder) and the receiver (decoder) under each fading channel is considered a separate task. For each fading channel realization, which will be equivalent to a sequence in the online scenario, a few samples corresponding to the current channel realization are transmitted to the receiver through the pilot signal. These samples sent via the pilot signal serve as the few-shot support and query sets used in the meta-learning algorithm. OML-CAE then utilizes the support and query sets of previous sequences accumulated in a task buffer as meta-training tasks to be used in a MAML algorithm. After meta-training, the MAML algorithm updates the initialization weight $\theta$ of the CAE. The initialization weight $\theta$ is then fine-tuned using the few samples in the current channel's support set, leading to the optimized weight $\theta^*$. $\theta^*$ is then used by the CAE until the next sequence arrives. It should be noted that during the first sequence, the Meta Training step is skipped as the task buffer is empty in the beginning. After the meta-training and finetuning steps, the received support and query sets are inserted in the task buffer to be used for meta-training in the next sequences. Task buffer size could vary depending on the application, the memory budget, and sample sizes. For example, in an IoT device with a limited memory budget, the task buffer could be lowered to increase memory efficiency.  

\section{Results}

\subsection{Experiment Settings}
In our experiments, we compared OML-CAE with three other baselines:
\begin{itemize}
    \item QPSK+MLE: A conventional communication system with QPSK modulation and maximum likelihood channel estimation and demodulation.
    \item CAE: a standard CAE trained from scratch for each new sequence using the few shot pilots available from the new channel realization.
    \item Joint-CAE: A CAE jointly trained on all previously observed channel realizations, and fine-tuned using the few shot samples from the new channel realization.
\end{itemize}
All methods use the same backbone CAE for fairness which is presented in Table \ref{tab:inner_model}. 

For OML-CAE's outer loop, we used 1 adaptation step, 0.0001 learning rate with Adam optimizer, 6000 iterations, 5 tasks per meta update, and a step scheduler with step size=300 and $\gamma$=0.9. For OML-CAE's inner loop, and the other methods, we used an SGD optimizer with a learning rate of 0.05 with 1000 iterations. The same hyperparameters used in OML-CAE's inner loop were used for training the other CAE baselines. It has to be noted since the CAE models will be trained using few-shot training datasets, and the used neural network backbone is relatively small, 1000 inner iterations do not pose a considerable training overhead. The task buffer size used for OML-CAE was set to 15. Our experiments were done under SNR=5 and SNR=10, and autoregressive Rayleigh fading process with $\rho=0.99$. Each experiment was performed using a sequence of 300 autoregressive Rayleigh fading channels.

\begin{table}[]
\caption{\small{Structure of the backbone Channel Autoencoder. k is the number of bits and $N_{ch}$ is the number of channel uses.}}
\renewcommand{\arraystretch}{1.4}
\resizebox{\columnwidth}{!}{\begin{tabular}{c|c|c|c|c|}
\cline{2-5}
                                                                                               & \textbf{Layer} & \textbf{Input}           & \textbf{Output}      & \textbf{Actsion Function} \\ \hline
\multicolumn{1}{|c|}{}                                                                         & Dense Layer    & $2^k$     & 256                  & Leaky ReLU                   \\ \cline{2-5} 
\multicolumn{1}{|c|}{\textbf{\begin{tabular}[c]{@{}c@{}}Encoder\\ (Transmitter)\end{tabular}}} & Dense Layer    & 256                      & 256                  & Leaky ReLU                   \\ \cline{2-5} 
\multicolumn{1}{|c|}{}                                                                         & Dense Layer    & 256                      & $2*{N_{ch}}$               & Leaky ReLU                   \\ \hline
\multicolumn{1}{|c|}{}                                                                         & Dense Layer    & $2*N_{ch}$ & 256                  & Leaky ReLU                   \\ \cline{2-5} 
\multicolumn{1}{|c|}{\textbf{\begin{tabular}[c]{@{}c@{}}Decoder\\ (Receiver)\end{tabular}}}    & Dense Layer    & 256                      & 256                  & Leaky ReLU                   \\ \cline{2-5} 
\multicolumn{1}{|c|}{}                                                                         & Dense Layer    & 256                      & 256                  & Leaky ReLU                   \\ \cline{2-5} 
\multicolumn{1}{|c|}{}                                                                         & Dense Layer    & 256                      & $2^k$ & Softmax                      \\ \hline
\end{tabular}}
\label{tab:inner_model}
\end{table}

\subsection{SER Experiment}

\begin{figure}
\vspace{-5pt}
\subfloat[][SNR 5]{\includegraphics[width=0.49\columnwidth]{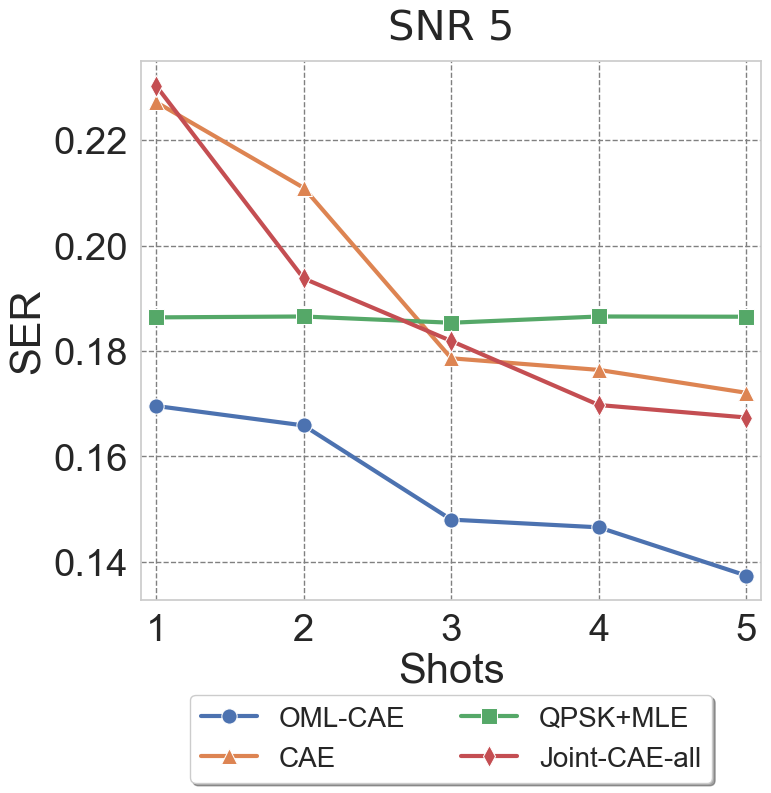}}
 \subfloat[][SNR 10]{\includegraphics[width=0.49\columnwidth]{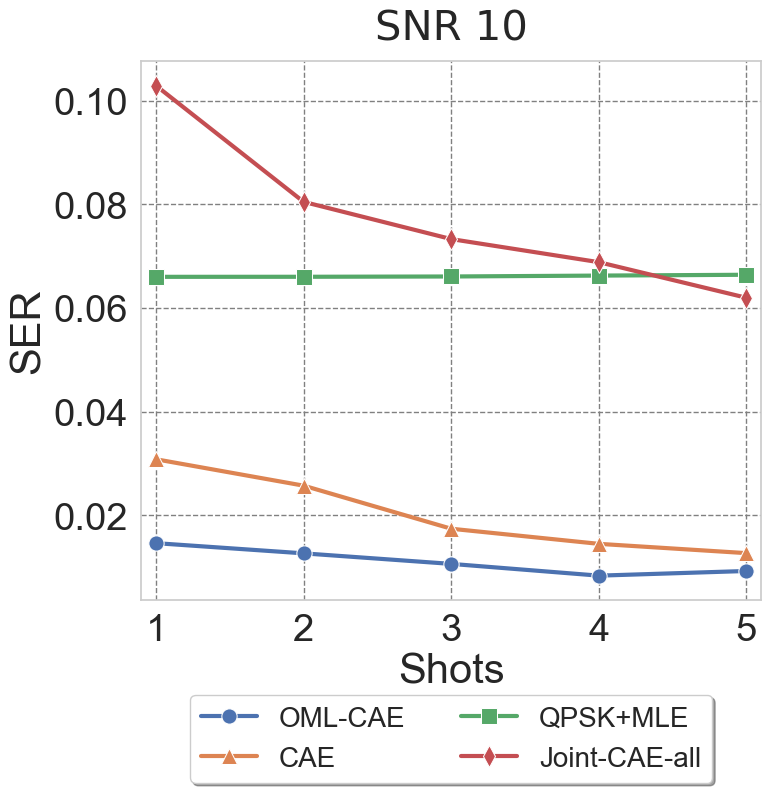}}

\caption{SER of the compared CAE methods in 1 to 5 shots. Subfigures (a) and (b) represent SNR=5 and SNR=10 respectively. Number of bits=4, Channel uses=2. For QPSK+MLE, each two bits are transmitted through different channels using QPKS and an error occurs if any of the 4 bits is demodulated incorrectly.}
\label{fig:4b2Nch}
\end{figure}

In a case of 4 bits and 2 channel uses, Figure \ref{fig:4b2Nch} presents SER (symbol error rate) of the CAE-based and the conventional baselines over the number of pilots for each symbol in SNR=5 and SNR=10. We can observe OML-CAE outperforms the other baselines in all cases. In SNR=5 the effectivenes of OML-CAE over the other baselines is more notable than in SNR=10, specially when compared to CAE.

As we are considering block fading, the only difference between each received pilot (i.e. training sample) for the same symbol, is the added noise to the sample when it is transmitted over the channel. Thus, in an ideal case of having no noise, one pilot per each symbol (i.e. one shot) would be enough information for the CAE to learn the channel and any more number pilots would be redundant. In realistic scenarios the noise is always present, but the lower the noise, the more information each individual pilot provides and there is less need for more number of pilots. To this end, SER improvement of OML-CAE will be more prominent in lower SNRs where there are only a few and less valuable pilots for training, as OML-CAE also utilizes the knowledge it has gained from previously observed channel realizations to adapt to the new channel. The comparatively smaller improvement of OML-CAE over CAE in SNR=10 can hence be interpreted this way. 

We can further observe that the Joint learning model has comparable results with CAE in SNR=5 and performs considerably worse than the other methods in SNR=10. Although the fading channels in different sequences are correlated, joint training on the corresponding pilots of previous channels does not provide useful information for new channels and further confuses the model in lower noise scenarios (SNR=10).

\begin{figure}
\subfloat[][SNR 5]{\includegraphics[width=0.49\columnwidth]{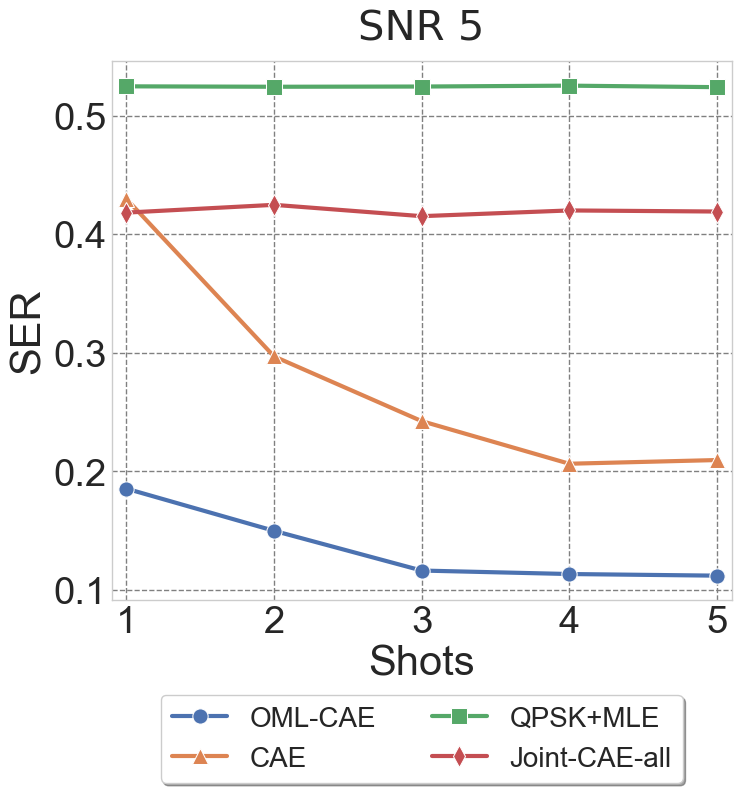}}
 \subfloat[][SNR 10]{\includegraphics[width=0.49\columnwidth]{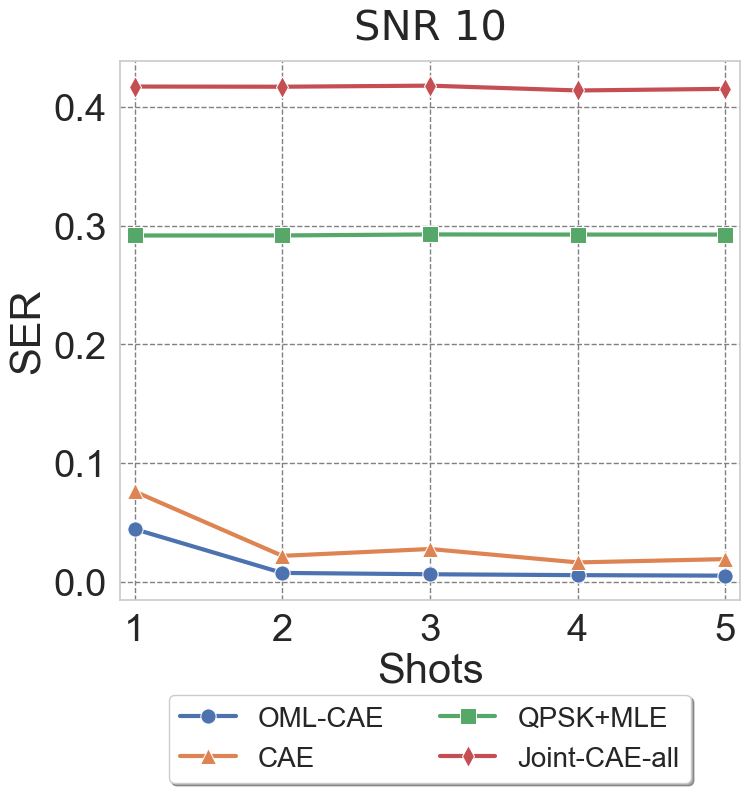}}

\caption{\small{SER of the compared CAE methods in 1 to 5 shots. Subfigures (a) and (b) represent SNR=5 and SNR=10 respectively. Number of bits=6, Channel uses=3. For QPSK+MLE, each two bits are transmitted through different channels using QPKS and an error occurs if any of the 6 bits is demodulated incorrectly.}} 
\label{fig:6b3Nch}
\vspace{-5pt}
\end{figure}

Figure \ref{fig:6b3Nch} provides the same results as Figure \ref{fig:4b2Nch}, but with 6 bits and 3 channel uses. Observations regarding OML-CAE are the same as Figure \ref{fig:4b2Nch}, with OML-CAE significantly outperforming the other methods in SNR=5, and moderately outperforming CAE in SNR=10. In the case Joint learning, it can be deducted that negative knowledge transfer from previously observed channels to the channel at hand has occurred. Although the different fading channels are correlated, it can be deducted they are different enough that the joint pre-training only results in confusion for the backbone CAE model. Moreover, other than pre-training the joint-CAE on all previous channels, we also tested its performance with limited buffer sizes of 5 and 15, meaning that the model was pre-trained on the last 5 or 15 previous channels. As the results remained nearly the same, we did not include them in Figures \ref{fig:4b2Nch} and \ref{fig:6b3Nch} to avoid confusing visualizations.

\begin{figure}[t!]
    \centering
    \begin{subfigure}[t]{0.48\columnwidth}
        \centering
        \includegraphics[height=1.2in]{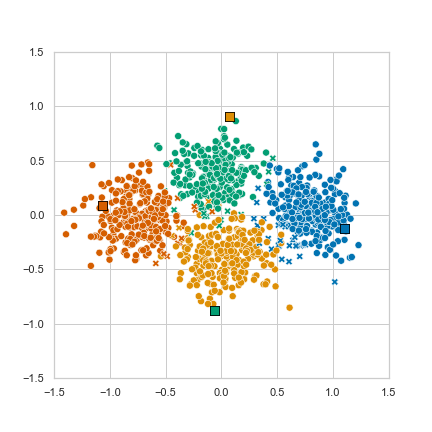}
        \caption{Sequence 1, OML-CAE}
    \end{subfigure}%
    ~ 
    \begin{subfigure}[t]{0.48\columnwidth}
        \centering
        \includegraphics[height=1.2in]{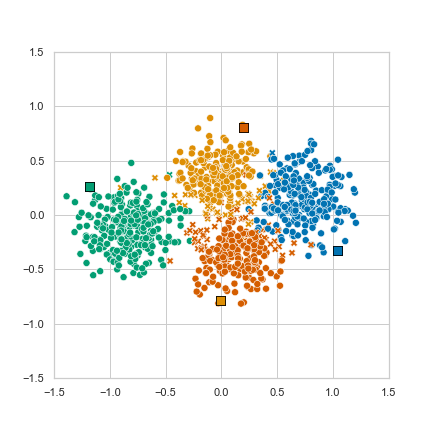}
        \caption{Sequence 1, CAE}
    \end{subfigure}
    
    \medskip

    \begin{subfigure}[t]{0.48\columnwidth}
        \centering
        \includegraphics[height=1.2in]{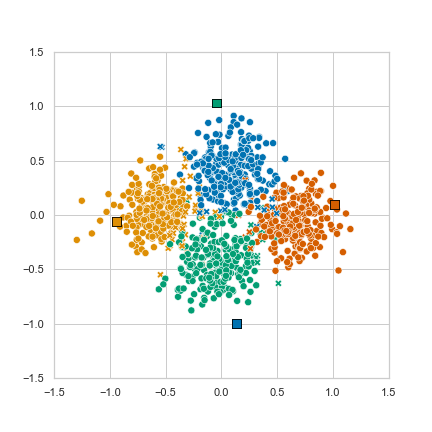}
        \caption{Sequence 2, OML-CAE}
    \end{subfigure}%
    ~ 
    \begin{subfigure}[t]{0.48\columnwidth}
        \centering
        \includegraphics[height=1.2in]{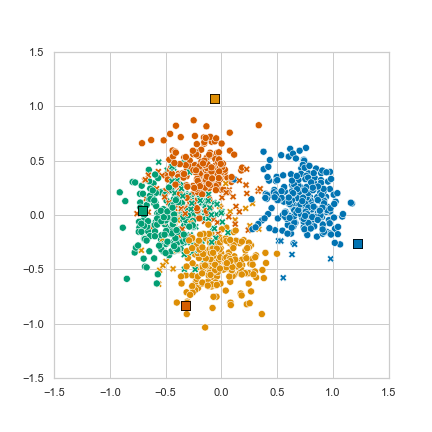}
        \caption{Sequence 2, CAE}
    \end{subfigure}

    \medskip

    \begin{subfigure}[t]{0.48\columnwidth}
        \centering
        \includegraphics[height=1.2in]{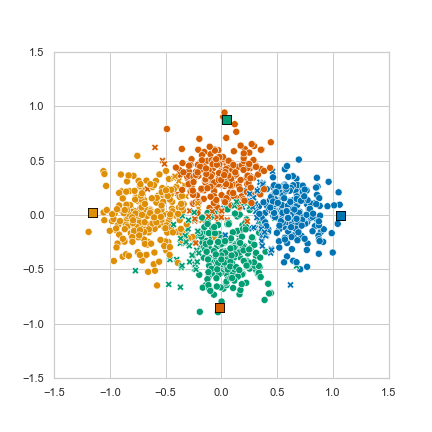}
        \caption{Sequence 3, OML-CAE}
    \end{subfigure}%
    ~ 
    \begin{subfigure}[t]{0.48\columnwidth}
        \centering
        \includegraphics[height=1.2in]{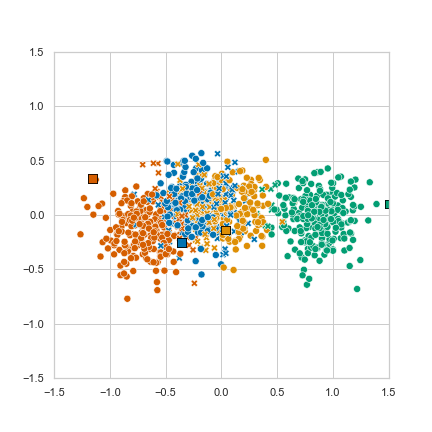}
        \caption{Sequence 3, CAE}
    \end{subfigure}

    \medskip

    \begin{subfigure}[t]{0.48\columnwidth}
        \centering
        \includegraphics[height=1.2in]{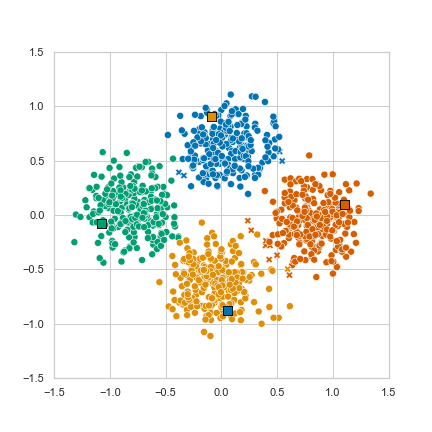}
        \caption{Sequence 4, OML-CAE}
    \end{subfigure}%
    ~ 
    \begin{subfigure}[t]{0.48\columnwidth}
        \centering
        \includegraphics[height=1.2in]{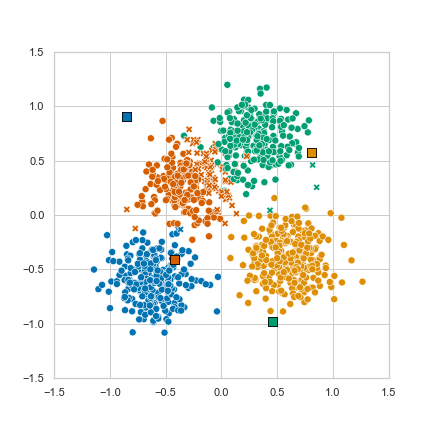}
        \caption{Sequence 4, CAE}
    \end{subfigure}
    \caption{\small{Received signals at the receiver for four subsequent Rayleigh fading channels with SNR=5 under OML-CAE (left) and CAE (right) in a case of 2 bits and 1 channel use and 5 shots. Marker colors indicate symbol label. Circle and cross markers indicate correct and wrong symbol detection respectively. Squares indicate the constellation designed by the encoder.}} 
    \label{fig:constellation_designs}
\end{figure}

We have also provided a visual comparison between OML-CAE and CAE in Figure \ref{fig:constellation_designs}. As the Joint-CAE method did not perform well, we have not included it in this figure. Figure \ref{fig:constellation_designs} provides visualizations of the received and transmitted symbols by OML-CAE and CAE under four subsequent autoregressive Rayleigh fading channels in SNR=5 with 5 shots. While it is evident that there is not one unique optimal modulation design for each channel, it is clear from the figures that the constellations designed by OML-CAE separate the symbols more effectively, leading to smaller SERs, and showing stronger adaptability to new fading channels.

\subsection{Pilot Efficiency Experiment}

We argued that if the CAE framework is going to be practically utilized in a dynamic scenario, it is reasonable to assume the CAE will encounter a few-shot learning scenario, where a small number of pilots would be available for the CAE to adapt to the channel. OML-CAE was proposed with this assumption and particularly with the claim that it allows lower SER in few-shot scenarios. From a different perspective, we can view this as a claim of improving efficiency. To reach a specific SER, OML-CAE requires less number of pilots to be available, allowing more room for actual data to be transmitted rather than pilots, and lowering the pilot overhead overall.

\begin{figure}[htp]
\vspace{-5pt}
\begin{center} 
\includegraphics[width=0.69\columnwidth]{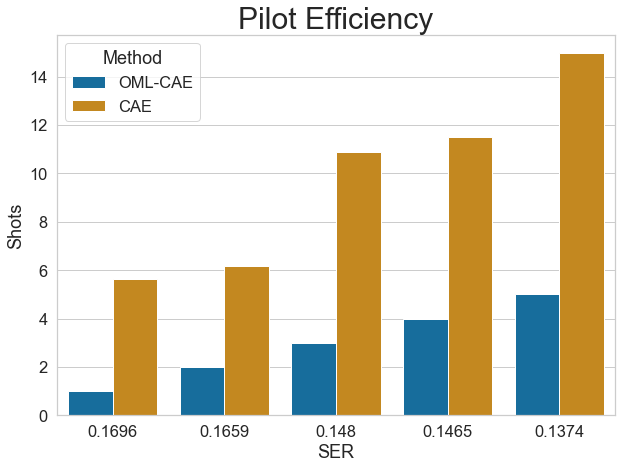}
\caption{\small{Pilot efficiency in terms of number of shots used in training tasks for OML-CAE and CAE to achieve the same SER. SNR=5, number of bits=4, number of channels=2.} }
\label{fig:efficiency} 
\end{center} 
\vspace{-5pt}
\end{figure}

Figure \ref{fig:efficiency} compares the number of shots required for OML-CAE and the CAE baseline to reach the same SER. SER values in the x-axis are the corresponding SERs achieved with OML-CAE in 1-shot to 5-shot scenarios. The number of shots is an integer in reality, but the values of CAE are not presented as integers. We performed an interpolation on CAE results from 1-shot to 30-shot scenarios, with the intention of being able to find the imaginary number of shots that would have been required if it were to reach the exact SERs that OML-CAE has reached. With CAE requiring roughly 3.8 times on average more number of pilots to reach the same SER as OML-CAE, we can observe OML-CAE significantly improves efficiency. It has to be noted that the conventional QPSK+MLE could not reach the SER rates achieved by OML-CAE even with large number of pilot samples, and thus, could not be compared to OML-CAE in terms of pilot efficiency.  

\section{Conclusion}

In this paper, we introduced OML-CAE, an online meta-learning framework for channel autoencoders with the aim of efficient adaptation to new fading channels in dynamic scenarios using few number of pilots. Through empirical evaluation, OML-CAE demonstrates superior performance in few-shot scenarios where the number of pilots is limited, particularly in lower SNR cases. Furthermore, the evaluation of efficiency highlights that OML-CAE achieves comparable SERs with substantially fewer shots compared to a conventional CAE, emphasizing its resource efficiency in dynamic communication environments. The higher efficiency of OML-CAE than conventional CAEs allows fewer pilots to be sent and consequently, enhances data rate by creating more bandwidth for the transmission of actual data.


\section{Acknowledgment}
This work is funded in part by the AFRL Summer Faculty Fellowship Program (SFFP) under contract number FA8750-20-3-1003.  
Any opinions, findings and conclusions, or recommendations expressed in this material are those of the authors and do not necessarily 
reflect the views of the U.S. Government.


{\tiny 
\printbibliography
}

\end{document}